\documentclass[10pt,twocolumn,letterpaper]{article}

\usepackage{iccv}
\usepackage{booktabs}
\usepackage[table,xcdraw]{xcolor}
\usepackage{times}
\usepackage{epsfig}
\usepackage{graphicx}
\usepackage{amsmath}
\usepackage{amssymb}
\usepackage{caption}
\usepackage{bm}
\usepackage{makecell}
\usepackage[ruled]{algorithm2e}
\SetKwInput{kwHypParam}{Hyper-parameter}
\usepackage{multirow}
\usepackage[symbol]{footmisc}

\usepackage[pagebackref=true,breaklinks=true,letterpaper=true,colorlinks,bookmarks=false]{hyperref}

\iccvfinalcopy 

\def\iccvPaperID{4730} 

\ificcvfinal\fi

\begin{document}


\title{PARF: Primitive-Aware Radiance Fusion for Indoor Scene Novel View Synthesis}





\author{Haiyang Ying$^1$,\ \ 
        Baowei Jiang$^1$,\ \ 
        Jinzhi Zhang$^1$,\ \ 
        Di Xu$^2$,\ \ 
        Tao Yu$^1$\footnotemark[2],\ \ 
        Qionghai Dai$^1$,\ \ 
        Lu Fang$^1$\footnotemark[2]
        \vspace{0.1cm}
    \and
        $^1$Tsinghua University, \ \ 
        $^2$Huawei Cloud 
}%



\twocolumn[{%
\renewcommand\twocolumn[1][]{#1}%
\maketitle
\ificcvfinal\thispagestyle{empty}\fi
\begin{center}
    \centering
    \captionsetup{type=figure}
    \includegraphics[width=1.0\textwidth]{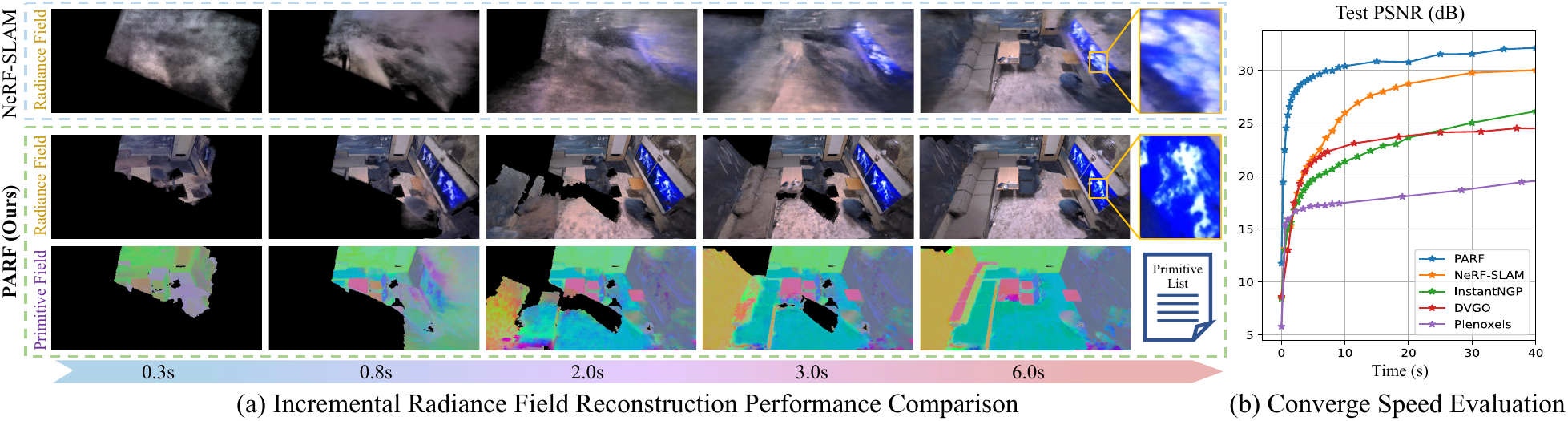}
    \captionof{figure}{
    Performance comparison with the state-of-the-art radiance field reconstruction methods on Replica dataset. 
    With the proposed hybrid representation and primitive-aware fusion framework, our method PARF enjoys significantly faster convergence and high-quality rendering for indoor scene novel view synthesis. 
    In (a) the incremental reconstruction setting, we assume a SLAM system with the tracking speed of 10fps. More resources can be found in our \href{https://oceanying.github.io/PARF/}{Project Page}.
    }
    \label{fig.teaser}
\end{center}%
}]

\footnotetext[2]{The corresponding authors are Lu Fang (fanglu@tsinghua.edu.cn, \href{http://www.luvision.net/}{http://www.luvision.net/}) and Tao Yu (ytrock@tsinghua.edu.cn).}

\begin{abstract}
This paper proposes a method for fast scene radiance field reconstruction with strong novel view synthesis performance and convenient scene editing functionality. The key idea is to fully utilize semantic parsing and primitive extraction for constraining and accelerating the radiance field reconstruction process. To fulfill this goal, a primitive-aware hybrid rendering strategy was proposed to enjoy the best of both volumetric and primitive rendering. We further contribute a reconstruction pipeline conducts primitive parsing and radiance field learning iteratively for each input frame which successfully fuses semantic, primitive, and radiance information into a single framework. Extensive evaluations demonstrate the fast reconstruction ability, high rendering quality, and convenient editing functionality of our method. 


\end{abstract}

\section{Introduction}

Indoor scene 3D reconstruction and novel view synthesis (NVS) is a long-lasting classical topic in the field of computer vision for decades, which is widely used in virtual reality, robot perception, and visualization.
Classic indoor scene reconstruction methods ~\cite{ORB-SLAM,KinectFusion} focus on geometric registration and fusion~\cite{KinectFusion} based on feature matching, bundle adjustment~\cite{dai2017bundlefusion}, and multi-view stereo~\cite{yao2018mvsnet} algorithms. However, these methods rely on discrete point clouds or voxels for scene representation, which results in high memory overhead and limited ability to describe scene details, making it challenging to achieve realistic NVS effects.

The emergence of implicit continuous representations based on neural networks has revolutionized 3D vision tasks. 
NeRF~\cite{martin2020nerf} represents the density and color fields of the scene using an implicit representation. 
Coupled with volume rendering techniques ~\cite{468400,10.1145/383507.383515,1196003}, NeRF achieves a simple but effective pipeline for end-to-end radiance field reconstruction. NeRF not only enables realistic novel view synthesis, but also facilitates 3D structure, material, and appearance recovery. 
However, NeRF-based methods tend to fit a diffused density field to the ground truth geometry surface for achieving view-dependent volume rendering effects, which may not be suitable for view extrapolation due to the lack of a sharp geometry constraint. 
Although incorporating depth information can constrain the learning of implicit geometry field, generating accurate samples for view extrapolation and achieving faster convergence remains challenging as NeRF requires relatively redundant sampling around the surface for volume rendering ~\cite{rosinol2022nerf}. 
Additionally, training NeRF in a manner of pixel-independent strategy neglects the global geometric consistency of the whole scene, which introduces noise and artifacts in the final reconstruction. 

To overcome this challenge, primitive-based methods such as NeurMiPs~\cite{lin2022neurmips} use global plane prior extracted from traditional primitive detection methods~\cite{pm-10.1007/978-3-642-33718-5_62,Holz2012RealTimePS,pm-Schnabel2007EfficientRF}. These methods typically use a fixed number of planes to fit the reconstructed point cloud obtained from other methods~\cite{pm-Attene2010HierarchicalSR,pm-Lafarge2012CreatingLC,pm-Rabbani2007AnIA}. 
This global structure prior effectively regularizes the implicit density field in the planar regions, thereby improving view extrapolation performance. 
However, for regions that are difficult to describe with planes, such as curved surfaces and thin structures, the boundary of the fitted plane suffers from obvious discontinuity artifacts. 


In this paper, we aim to establish an incremental radiance field reconstruction pipeline based on NeRF and semantic parsing for much higher performance, no matter view interpolation or extrapolation, with an order of magnitude fewer training iterations than SOTA methods. 
Our key innovation is a divide-and-conquer strategy that makes the representation ultra-simple in global primitive regions while keeping it complex in non-plane local details.

In light of this, we propose \textbf{P}rimitive-\textbf{A}ware \textbf{R}adiance \textbf{F}usion, named \textbf{PARF}, for indoor scene novel view synthesis. 
Our key idea is that: Indoor scene always contains many planar regions, and by leveraging the global primitive prior of planar regions and the local implicit representation for non-planar regions, we can achieve much better performance with strong semantic guidance. 
However, representing, fusing, and training both primitive and non-primitive representation in the same radiance field from sequential RGB-D inputs in real-time is non-trivial. 
In order to solve the problems above, PARF proposes a hybrid representation that uses discrete semantic volume as a medium to integrate planar semantics into the continuous and implicit scene radiance field. 
This allows for a primitive-aware sampling process in volume rendering, resulting in improved efficiency and quality. 
Additionally, PARF dynamically maintains a global scene plane representation and can fuse and differentiate planar regions through dynamic fusion and adaptive update. This enables efficient and noise-robust optimization of the radiance field, as well as direct semantic editing capabilities.
Overall, PARF successfully incorporates semantic parsing and primitive merging into a radiance fusion framework, enabling efficient training, high-quality rendering, and semantic editing.

The contributions of PARF can be summarized as: 

\begin{itemize}
    \item We propose PARF, a novel hybrid scene representation to decompose the radiance field into primitive-based and volume-based components in a unified form. 

    \item We contribute an incremental reconstruction framework for primitive-aware radiance fusion, which effectively leverages the benefits of semantic parsing, primitive merging, and neural representation for indoor scene reconstruction.

    \item Extensive evaluations demonstrate that our method enjoys fast convergence, robust view extrapolation performance, and convenient scene editing ability. 
    
\end{itemize}

\section{Related Works}
\subsection{Neural Implicit Rendering and Fusion}
Neural Radiance Field~\cite{mildenhall2020nerf, martin2020nerf, zhang2020nerf++, wang2021ibrnet, liu2020neural, muller2022instant} is an approach that utilizes coordinate-based MLP as implicit scene representations to continuously encode scene geometry, which achieves high-quality and view-dependent appearance modeling. Signed Distance Field is also an implicit representation that is beneficial to model a continuous geometric surface~\cite{DeepSDF, azinovic2022neuralrgbd, zhu2022niceslam, guo2022manhattan, yu2022monosdf}.

Towards indoor scene fusion, NeuralRGBD~\cite{azinovic2022neuralrgbd} models and optimizes the scene geometry as a continuous SDF function and achieves high completeness though the training time is quite long. NICE-SLAM~\cite{zhu2022niceslam} proposes a dense SLAM system that optimizes a hierarchical representation with pre-trained geometric priors which, enables detailed reconstruction on large indoor scenes. NeuralRecon~\cite{sun2021neuralrecon} establishes a learning-based TSDF fusion module based on GRU to guide the network to fuse features from previous fragments in real time. 

For indoor scene rendering, NeRFusion~\cite{zhang2022nerfusion} applies a pre-trained fusion model for real-time RGB radiance fusion for novel view synthesis.
Based on InstantNGP~\cite{muller2022instant}, NeRF-SLAM~\cite{rosinol2022nerf} create a NeRF-based SLAM system with extra depth as supervision signal to achieve real-time radiance field reconstruction. 

However, a limitation of volumetric representations is that the optimization is applied on the integral of the radiance field without sufficient prior. This can lead to biased and inconsistent geometry and therefore results in bad view extrapolation and slow convergence speed. Though prior-based methods like ~\cite{niemeyer2022regnerf, kim2022infonerf} uses strong regularization on visual patches show satisfactory results under sparse-view setting, the performance gap still remains between the controlled scenes with structured observations and real-world scenes with unstructured captures.

\subsection{Primitive based Rendering and Fusion}

Structural scene prior has been proven to be beneficial in neural rendering and fusion~\cite{popovic2022manhhatan_unknown, lin2022neurmips, guo2022manhattan, chen2022mobilenerf, xie2022planarrecon, ying2022parsemvs}. 

ManhattanSDF~\cite{guo2022manhattan} uses the Manhattan prior to constrain the normal of an implicit SDF field, which highly relies on known semantics of scene partition and the Manhattan frame. Further work~\cite{popovic2022manhhatan_unknown} employs self-supervision of depth and normals through the Manhattan prior and volumetric rendering without the Manhattan frame. But these two works are still limited to the Manhattan assumption, which is not sufficient for modeling unordered planes in 3D space.

To handle planes with free poses, PlanarRecon~\cite{xie2022planarrecon} proposes to fusion bounded planes in an incremental manner based on NeuralRecon~\cite{sun2021neuralrecon} but suffers from incomplete fusion results. NeurMiPs~\cite{lin2022neurmips} uses SFM point cloud to decompose the scene into optimizable planar experts, which benefits from fast planar rendering and optimization.
However, purely plane-based modeling may lead to difficulty in complex scene modeling, especially when observation is insufficient. 

From the view of rendering efficiency, MobileNeRF~\cite{chen2022mobilenerf} decomposes the scene into a set of polygons with textures representing binary opacities and feature vectors. However, since the triangle primitive is quite small, it still suffers from overfitting and cannot handle view extrapolation.

\section{Representation}
We present a novel primitive-aware hybrid representation to model the scene in a hybrid manner. Based on a primitive-aware semantic volume, the scene can be divided into volume-based and primitive-based regions automatically. 
Both of dense volume rendering and primitive-based rendering can be applied via a unified representation.
In this section, we will first recap the NeRF-based volume rendering in Sec.~\ref{sec:volume_based_rendering}, and introduce the primitive-based rendering method in Sec.~\ref{sec:primitive_based_rendering}. Then the core idea, primitive-aware hybrid representation, will be introduced in detail in Sec.~\ref{sec:primitive_aware_hybrid_representation}.

\begin{figure*}[t] 
\centering
    \includegraphics[width=0.8\linewidth]{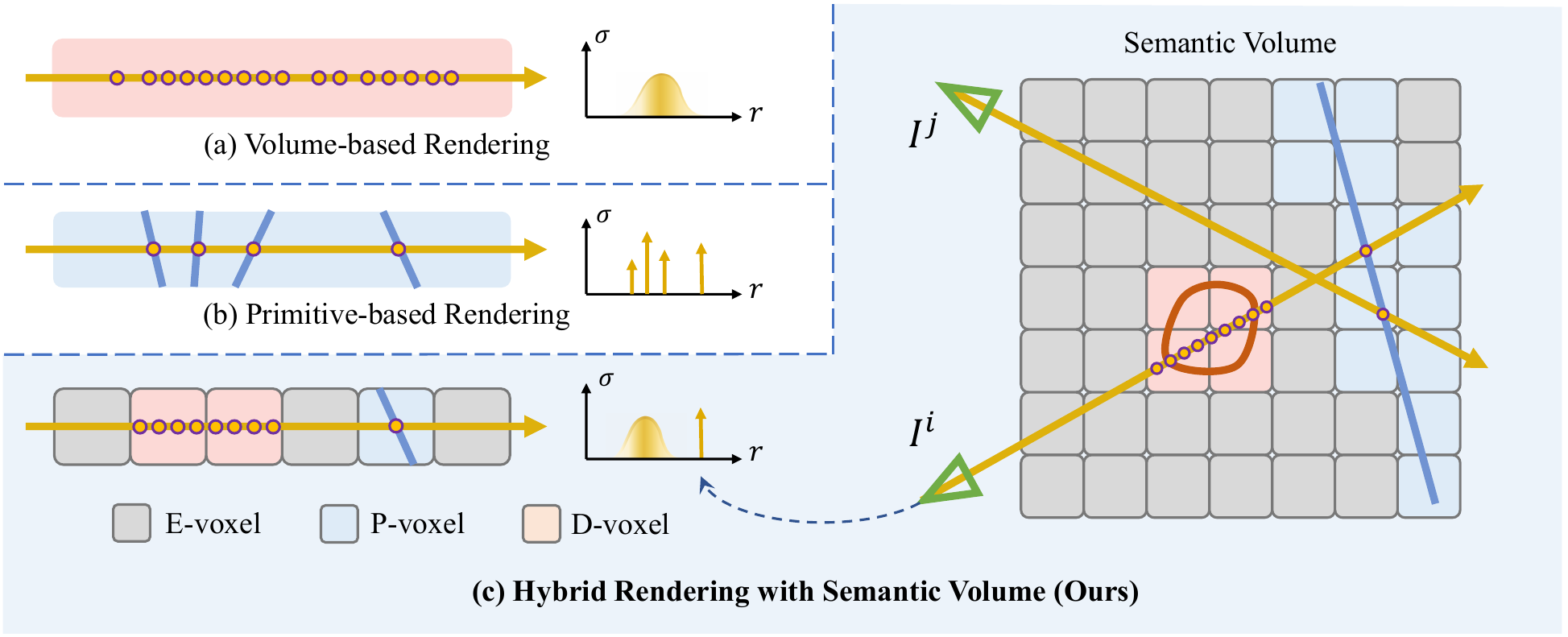} 
    \caption{Representation. By discretizing the scene with a semantic volume, the proposed (c) \textbf{primitive-aware hybrid rendering} enables highly efficient sampling and rendering in the mixture of both (a) volumetric and (b) primitive rendering.}\label{fig:PARF_representation}
\end{figure*}

\subsection{Volume-based Rendering}\label{sec:volume_based_rendering}
We utilize the radiance field~\cite{mildenhall2020nerf} as the basis of our representation. More specifically, given the position $\mathbf{x}_i \in \mathbb{R}^3$ and the view direction $\mathbf{d}_i \in \mathbb{R}^2$, an MLP network $F_{\mathbf{\Theta}}$ will act as a decoder and output the per-point attributes:
\begin{equation}
    \sigma_i = F_{\mathbf{\Theta}}(\gamma(\mathbf{x}_i)), 
    \ \ \ 
    \mathbf{c}_i = F_{\mathbf{\Theta}}(\gamma(\mathbf{x}_i), SH(\mathbf{d}_i)),
    \label{eq:nerf_mlp}
\end{equation}
where $\sigma_i \in \mathbb{R}$ is the view-independent density and  $\mathbf{c}_i \in \mathbb{R}^3$ is the view-dependent RGB color. 
$\gamma(\cdot)$ and $SH(\cdot)$ are positional encoding functions based on multi-resolution hashing~\cite{muller2022instant} and spherical harmonics respectively.
In order to model the semantic information of the space additionally, inspired by semanticNeRF~\cite{zhi2021place}, we add a semantic head to the MLP $F_{\mathbf{\Theta}}$ and get the per-point semantic information $\mathbf{s}_i \in \mathbb{R}^4$: $\mathbf{s}_i = F_{\mathbf{\Theta}}(\gamma(\mathbf{x}_i))$,
where $\mathbf{s}_i=(\mathbf{n}_p, d_p)$ indicates the primitive the queried point $\mathbf{x}_i$ is located on. We define each primitive as a 3D plane which will be introduced in Sec.~\ref{sec:primitive_based_rendering}. Instead of predicting the discrete object class labels~\cite{zhi2021place}, our semantic logits $\mathbf{s}_i$ are continuous and indicate geometric-level semantic information of the scene.

Then color and density will be integrated along the ray to get the rendered pixel color $\mathbf{c}(\mathbf{r})$.
\begin{equation}
    \mathbf{c}(\mathbf{r})=\sum_{i=1}^{N} T_i\alpha_i \mathbf{c}_i, 
    \ \ \ 
    T_i=\prod_{j=1}^{i-1} (1-\alpha_i)
    \label{eq:volume_rendering}
\end{equation}
where $\alpha_i = 1-exp(-\sigma_i\delta_i)$ is the opacity and $\delta_i= r_{i+1} - r_i$ is the distance between adjacent samples. Besides RGB color, Eq.~\ref{eq:volume_rendering} can also be used to render depth and semantic values as:
\begin{equation}
    d(\mathbf{r})=\sum_{i=1}^{N} T_i\alpha_i r_i, 
    \ \ \ 
    \mathbf{s}(\mathbf{r})=\sum_{i=1}^{N} T_i\alpha_i \mathbf{s}_i . 
    \label{eq:volume_rendering_2}
\end{equation}

\subsection{Primitive-based Rendering}\label{sec:primitive_based_rendering}
Since dense volume rendering suffers from expensive sampling and ambiguous geometry around the ground truth surface, geometric primitive-based rendering may be an alternative choice.
We define each primitive as a 3D plane $\mathbf{p} = \{\mathbf{n}_p, d_p\}$, where $\mathbf{n}_p \in \mathbb{R}^3$ is the plane normal and $d_p \in \mathbb{R}_+$ is the distance from the origin point to the plane, Each primitive $\mathbf{p} = \{\mathbf{n}_p, d_p\}$ holds $\mathbf{n}_p \cdot \mathbf{x} = d_p$ for the point $\mathbf{x}$ located on it. 

Given a ray $\mathbf{r}=\{ \mathbf{o}, \mathbf{d} \}$ and a primitive $\mathbf{p} = \{\mathbf{n}_p, d_p\}$, the ray-primitive intersection point can be calculated analytically:
\begin{equation}
    \mathbf{x}=\mathbf{o} + \frac{\mathbf{d}_p-\mathbf{o}\cdot \mathbf{n}_p}{\mathbf{d}\cdot \mathbf{n}_p} \mathbf{d} . 
    \label{eq:ray_plane_intersect}
\end{equation}

We model primitives as colored and translucent planes so the same rendering method (Eq.~\ref{eq:nerf_mlp}-Eq.~\ref{eq:volume_rendering_2}) can be applied. 
However, the primitive intersections are often sparse and unevenly spaced, so $\delta_i= r_{i+1} - r_i$ is unreasonable in primitive-based rendering. To solve this, we assume each plane shares an equal and fixed thickness, i.e., $\delta_i = \delta_p$.

The primitive-based rendering is as follows: shooting a ray $\mathbf{r}=\{ \mathbf{o}, \mathbf{d} \}$ from the camera optical center to the space, computing all intersections with existing primitives $ \mathbf{P}_G $, sorting intersections $\{ \mathbf{x}_i \}$ by the distance from ray origin, sending intersection positions and ray directions into the MLP $F_\mathbf{\Theta}$ and executing the volume rendering as Eq.~\ref{eq:nerf_mlp} and Eq.~\ref{eq:volume_rendering} with fixed thickness $\delta_i = \delta_p$.



Though plane-based methods are very efficient~\cite{chen2022mobilenerf, lin2022neurmips}, pure primitive-based modeling may still lead to wrong geometry and blurry rendering results when observations are limited. On the other hand, volume-based rendering can relieve the problem by dense sampling for regions with complex geometry. In light of this, a hybrid representation that combines both primitive and non-primitive based rendering may help improve the fidelity of scene modeling and rendering results.

\subsection{Primitive-aware hybrid representation}\label{sec:primitive_aware_hybrid_representation}

Taking advantage of both volume rendering and primitive-based rendering, we design a hybrid voxel-based representation to achieve complex scene modeling and fast inference simultaneously, which also helps lift up convergence speed.

The basic idea is to create an indicator to help tell apart primitive and non-primitive regions.
Specifically, to represent a scene in a hybrid manner, we establish a dense semantic volume $\mathbf{V_s}$ accompanied by a list of primitive parameters $\mathbf{P}_G$ to describe the global semantic information of each point in the scene. Each voxel of the volume contains an integer label $v_{i} \in \mathbb{Z}, (v_{i} \ge -1)$ indicating the type of the voxel. Different sampling and rendering strategies are used for different types of voxels. We first apply ray marching in the semantic volume $\mathbf{V_s}$ to sample points, and a hybrid rendering method is utilized to render per-pixel color, depth, and semantic values to form the rendered images.

\noindent \textbf{Ray Marching}. To render a ray, we apply ray marching in the semantic volume to sample points for rendering. At each marching step, we determine which voxel the current point belongs to. Then the semantic label of the current voxel is checked by the semantic volume, and the label determines the sample operation we will execute. We define the following three kinds of voxel to guide the sampling:

\textit{ \textbf{E-voxel}} holds $v_{i}=-1$, which means the voxel is empty, and the marching process will skip this voxel without sampling.

\textit{ \textbf{D-voxel}} holds $v_{i}=0$, which means the voxel is occupied. Samples in this voxel will be dense and evenly spaced.

\textit{ \textbf{P-voxel}} has $v_{i} \ge 1$, which means the voxel is also occupied, but we apply primitive-based sampling and rendering in these voxels. Each $v_{i} \ge 1$ corresponds to one primitive in the maintained parameter list $\mathbf{P}_G$. We extract the parameter $\mathbf{p} = \{\mathbf{n}_p, d_p\}$ of the indicated primitive $v_{i}$ from $\mathbf{P}_G$. Then the ray-plane intersection point is calculated with Eq.~\ref{eq:ray_plane_intersect}, and its coordinate is saved for later rendering. After that, we set the coordinates of the next marching point as a point at a fixed distance $\psi$ behind the plane and continue the marching process until the sample point moves outside the semantic volume $\mathbf{V_s}$ or the ray has reached the maximum number of sampling steps. The fixed distance $\psi$ is set as the diagonal size of one voxel along the plane normal. 
%

\noindent \textbf{Hybrid Rendering}. After ray marching, we have gathered sampled point set $\{\mathbf{x}\}$ from the traversed D-voxels and P-voxels. Then the point set is sent into MLP $F_{\mathbf{\Theta}}$ to infer color and density. When calculating the opacity $\alpha_i$ of each point, we choose the thickness as $\delta_i=r_{i+1}-r_{i}$ for points sampled in D-voxels and $\delta_i=\delta_p$ for points sampled in P-voxels ($\delta_p$ is set to $1.0$ in all the experiments). Then the  pixel color can be rendered by Eq.~\ref{eq:volume_rendering}

One advantage of this hybrid semantic volume is that parametric parameters and boundaries of primitives are encoded into the scene in a unified manner, which means no extra plane parameterization is needed for each primitive.

\begin{figure*}[t] 
\centering
    \includegraphics[width=0.9\linewidth]{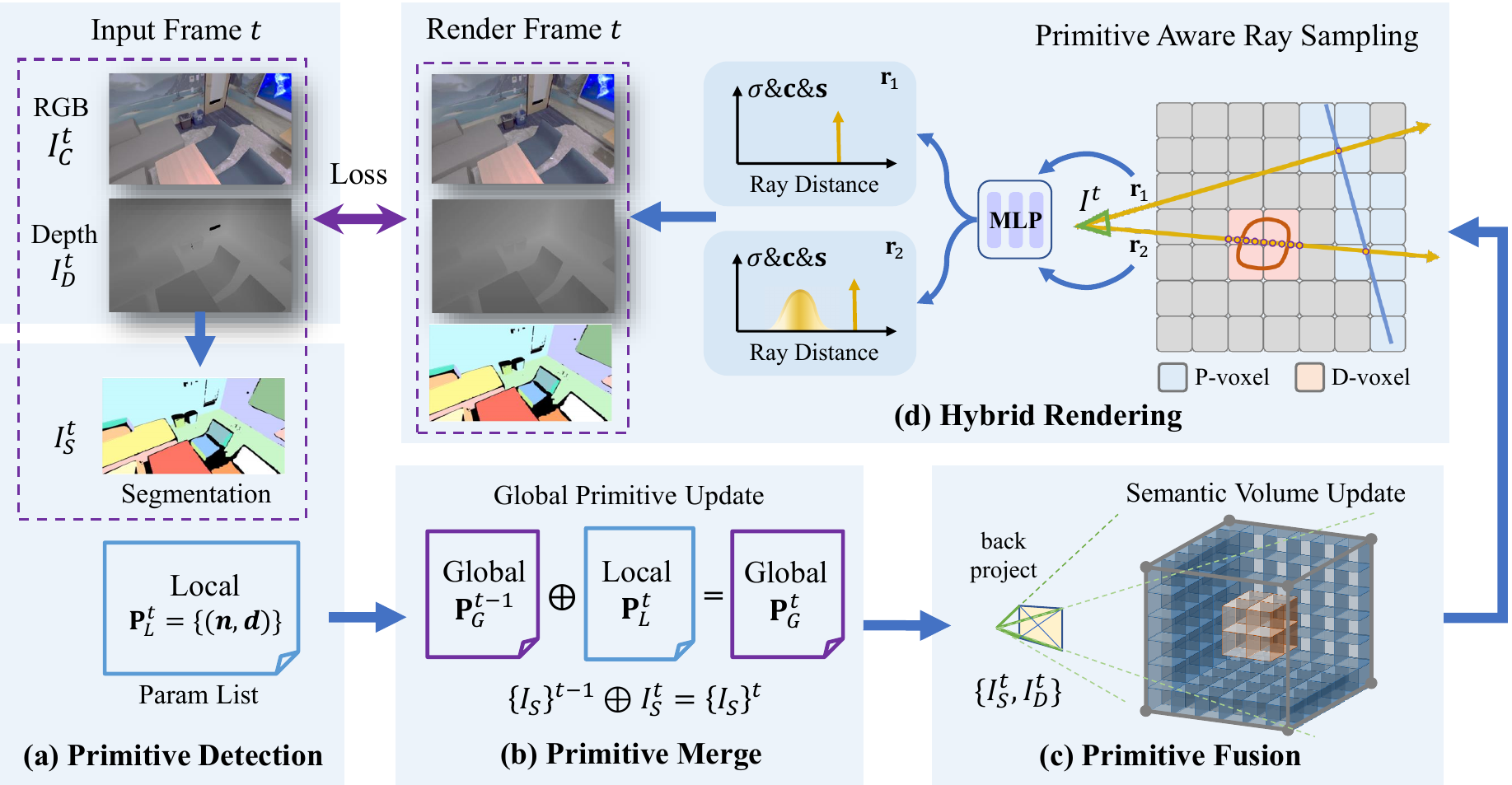} 
    \caption{Framework of PARF. For each input RGB-D frame, we apply plane detection to get semantic image $I_S^t$ and a local parameter list $\mathbf{P}_L^t$, which are then merged into the global primitive list. Then the depth frame $I_D^t$ and the updated semantic frame $I_S^t$ are fused into the semantic volume $\mathbf{V_s}$ to update the global representation. Through hybrid rendering, the color, depth, and semantic images can be rendered and supervised with input images and detected primitive images.}\label{fig:framework}
\end{figure*}

\section{Primitive-aware radiance fusion}\label{sec:primitive_aware_radiance_fusion}
Given a posed RGB-D sequence as input, we reconstruct a primitive-aware volumetric field for novel view synthesis in an incremental manner. We apply a plane detection algorithm for each input depth image to estimate plane parameters and merge them into the global plane list. After that, the new primitives will be fused into the semantic volume $\mathbf{V_s}$.
Finally, the MLP $F_{\mathbf{\Theta}}$ and the semantic volume $\mathbf{V_s}$ will be optimized together via the proposed hybrid rendering.


\subsection{Parametric primitive extraction}
Though depth sensors may give noisy observation, regions with continuous depth values provide strong prior for the existence of smooth surfaces. This prior is especially valuable for recovering texture-less regions, which is difficult for NeRF~\cite{mildenhall2020nerf} and MVS~\cite{schonberger2016colmap} methods. 
We choose plane as a similar prior to depict the low-level semantics of the scene. Inspired by TSDF Fusion, we propose to detect and fuse the semantic information into the semantic volume in an incremental manner. 

\noindent \textbf{Primitive detection}.
  Given depth frame $I^{t}_{D}$ at time $t$, a real-time plane detection method CAPE~\cite{proencca2018fast} is utilized to detect planes and get a parameter list $\mathbf{P}_L^t$ as well as a semantic image $I^{t}_{S}$, where each pixel with non-zero value corresponds to a plane in the list $\mathbf{P}_L^t$. 
  For each plane in $\mathbf{P}_L^t$, we validate its flatness by back-projecting the pixels to 3D space and calculating the mean point-to-plane error. If the error exceeds the threshold $\epsilon_1$, the plane will be refused, and the corresponding pixel value on $I^{t}_{S}$ will be set to zero.

			
				
			
			
  
\noindent \textbf{Primitive merge}.
 After plane detection, the detected planes $\mathbf{P}_L^t$ are compared and merged into the global plane list $\mathbf{P}_G^{t-1}$ to get $\mathbf{P}_G^t$. 
 For each plane $\mathbf{p}^t_j \in \mathbf{P}_L^t, j\in \left[ 1, J \right]$, the distance between $\mathbf{p}^t_j$ and each plane $\mathbf{p}_m \in \mathbf{P}_G^{t-1}, m\in\left[ 1,M \right]$ is calculated as:
 \begin{equation}
     dist(\mathbf{p}^t_j, \mathbf{p}_m) = \left | d^t_j\mathbf{n}^t_j - d_m\mathbf{n}_m \right | . 
 \end{equation}
 If all the distances are larger than threshold $\epsilon_2$, plane $\mathbf{p}^t_j$ will be added to $\mathbf{P}_G^{t-1}$ as a new plane. Otherwise, the pixels in $I^{t}_{S}$ correspond to plane $\mathbf{p}^t_j$ will be replaced with the index of the closest plane $\mathop{\arg\min}\limits_{m} dist(\mathbf{p}^t_j, \mathbf{p}_m)$.

 After merging $\mathbf{P}_L^t$ into the global list $\mathbf{P}_G^{t-1}$, we apply PCA to evaluate the normal $\widetilde{\mathbf{n}}_m$ of each plane $\mathbf{p}_m \in \mathbf{P}_G^t$ by sampling points in the last $t$ semantic index images. If $ \left | \widetilde{\mathbf{n}}_m - \mathbf{n}_m \right | > \epsilon_3$, then plane $\mathbf{p}_m$ will be removed from $\mathbf{P}_G^t$.
 In our experiments, the threshold values are $\epsilon_1=0.005$, $\epsilon_2=0.01$, and $\epsilon_3=0.1$.

\subsection{Primitive fusion in semantic volume}

 Inspired by TSDF Fusion based reconstruction, we further fuse the current semantic frame $I^{t}_{S}$ and the depth frame $I^{t}_{D}$ into the semantic volume for hybrid rendering. 
 
 At beginning of the fusion, we assign all voxels in semantic volume $\mathbf{V_s}$ as E-voxels ($v_i=-1$).
 When a new frame comes, we project all the grid points $\{x_i\}$ (center points of voxels) onto the current frame $t$ to get the pixel coordinates $\{u_i\}$ and the observed depth values $\{D^t(x_i)\}$. 
 
 If the projected semantic value $I^{t}_{S}(u_i)=0$,  which indicates non-primitive, then we apply a bilateral truncated band $B_1$  to threshold the valid grid points according to the depth value $I^{t}_{D}(u_i)$ of the projected pixel. The valid voxels will be assigned as D-voxels ($v_i=0$):
 \begin{equation}
     V^t_{v=0} = \{v_i | I^{t}_{D}(u_i) - B_1 < D^t(x_i) < I^{t}_{D}(u_i) + B_1\} .
 \end{equation}
 
 If $I^{t}_{S}(u_i)>0$, the pixel indicates a primitive $m \in \left[1,M \right]$. Firstly, we compute the ray-plane intersection with Eq.~\ref{eq:ray_plane_intersect} and get the observed depth values $\{S^t(x_i)\}$ of the intersection point for the following threshold operations. Secondly, we use a narrower bilateral band $B_2$ to threshold the voxels and assign plane index $m$ to these voxels as P-voxels. Thirdly, we take plane primitive as a strong regularization of the space, where there should be no occupied voxels between the camera $t$ and the observed plane $\mathbf{p}_m$. So we apply a unilateral truncation band $B_1$ to assign voxels behind the plane as D-voxels only, and the voxels located before the plane will be set to E-voxels.
\begin{equation}
     V^t_{v>0} = \{v_i | I^{t}_{D}(u_i) - B_2 < S^t(x_i) < I^{t}_{D}(u_i) + B_2\} , 
\end{equation}
 \begin{equation}
     V^t_{v=0} = \{v_i | I^{t}_{D}(u_i) + B_2 < S^t(x_i) < I^{t}_{D}(u_i) + B_1\},
 \end{equation}
 \begin{equation}
     V^t_{v=-1} = \{v_i | S^t(x_i) < I^{t}_{D}(u_i) - B_2\} . 
 \end{equation}
The bandwidth $B_1 = 6\psi, B_2=\psi$, where $\psi$ is the diagonal size of one voxel. Please refer to our supplementary material for more details.




This primitive-based fusion operation helps sparsify the space, which is beneficial for fast convergence. After the parametric semantic fusion, the updated semantic volume $\mathbf{V_s}$ can be used to execute hybrid rendering (Sec.~\ref{sec:primitive_aware_hybrid_representation}) and further optimization (Sec.~\ref{sec:optimization}).

\subsection{Implementation details}

\subsubsection{Optimization}\label{sec:optimization}

During training, we optimize the MLP $F_\mathbf{\Theta}$ and the semantic volume $\mathbf{V_s}$ via hybrid volume rendering. For volume rendering, we apply four loss functions: $\mathcal{L}_c = \sum_{\mathbf{r}}\left \| \mathbf{c}(\mathbf{r})-\mathbf{c}_{\mathrm{gt}}(\mathbf{r}) \right \|^2_2$, $\mathcal{L}_d = \sum_{\mathbf{r}}\left \| d(\mathbf{r})-d_{\mathrm{gt}}(\mathbf{r}) \right \|^2_2$, $\mathcal{L}_s = \sum_{\mathbf{r}}\left \| \mathbf{s}(\mathbf{r})-\mathbf{s}_{\mathrm{gt}}(\mathbf{r}) \right \|^2_2$, and $\mathcal{L}_{reg} = \sum_{\mathbf{r}} -o(\mathbf{r})\log(o(\mathbf{r}))$,
where $o(\mathbf{r})=\sum_{i=1}^{N} T_i\alpha_i$ is the opacity of each ray. $\mathcal{L}_{reg}$ is used to regularize each ray to be completely saturated or unsaturated. The total loss is: 
\begin{equation}
    \mathcal{L}_{total}= \mathcal{L}_{c} + \lambda_1\mathcal{L}_{d} + \lambda_2\mathcal{L}_{s} + \lambda_3\mathcal{L}_{reg} , 
\end{equation}
The hyper-parameters are set as $\lambda_1 = 1.0, \lambda_2=0.04, \lambda_3=0.001$ for all the experiments. 
With a cosine annealing schedule, the learning rate is set from $1e^{-2}$ to $3e^{-4}$. The ray number of each batch is $8192$, and each epoch contains $1000$ iterations. We train PARF for $5$ epochs for each scene.
%
We apply the same pruning strategy as InstantNGP~\cite{muller2022instant} to prune voxels with low density periodically, which helps sparsify the space and accelerate the optimization speed.

\subsubsection{Scene Editing}

The hybrid scene representation helps to achieve more convenient scene editing with the following actions.

\textit {{Primitive Deletion}}. Since each primitive holds a unique label $v_i$ in the semantic volume $\mathbf{V_s}$, the primitive can be easily removed by setting P-voxels labeled with $v_i \ge 1$ to E-voxels $v_i = -1$. 

\begin{figure}[htb] 
\centering
    \includegraphics[width=1.0\linewidth]{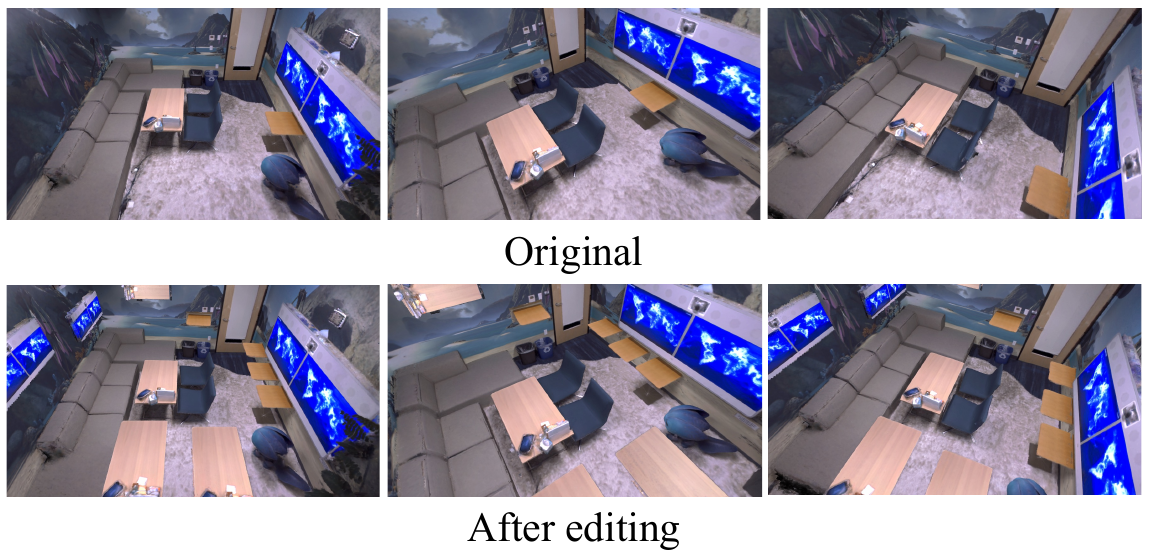} 
    \caption{Realistic semantic scene editing results.}
    \label{fig:editing}
\end{figure}

 \textit{Primitive Transformation}. 
 To transform primitives, We set up an extra editing volume $\mathbf{V_e}$ and a list $\mathbf{T_e}$ to store the editing operations.
Non-zero voxel $v^e_i \in \mathbf{V_e}$ indicates an editing operation $\mathbf{t}^e_i \in \mathbf{T_e}$.
During ray marching, if the ray arrives a voxel with $v^e_i \ge 1$ in $\mathbf{V_e}$, the current marching point $\{\mathbf{x}, \mathbf{d}\}$ will be transformed to $(\mathbf{x', d'}) = \mathbf{t}^e_i(\mathbf{x, d})$ according to $\mathbf{t}^e_i$. Then the transformed $(\mathbf{x', d'})$ will be sent into MLP $F_\mathbf{\Theta}$ for attributes inferencing and rendering.
%
The visualization of editing results are shown in Fig.~\ref{fig:editing}.


\begin{figure*}[htb]
  \centering
   \includegraphics[width=1.0\linewidth]{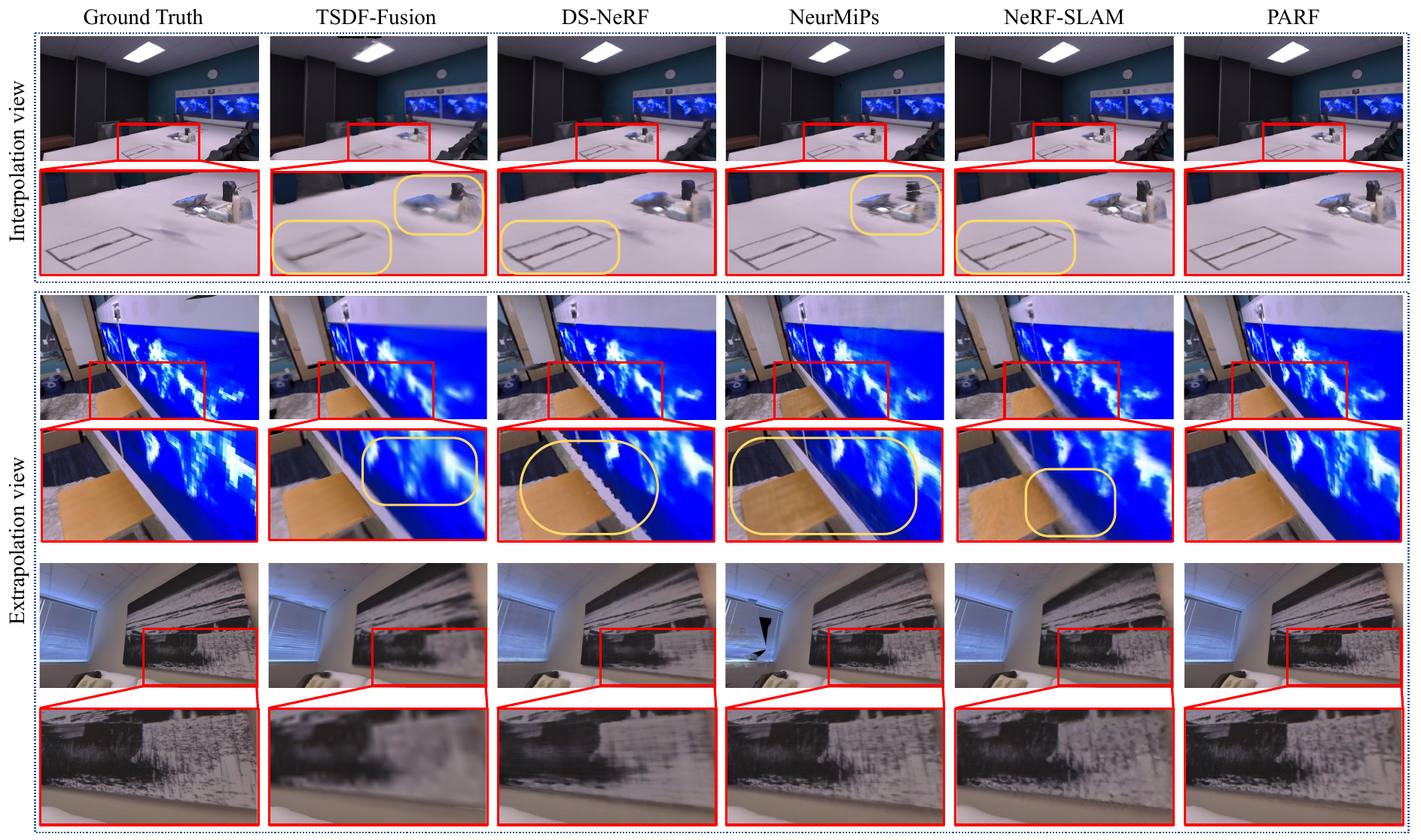}
   \caption{Qualitative comparison on both interpolation and extrapolation views of the Replica dataset. PARF shows high-quality rendering results under both settings, while other methods suffer from blurry patterns, geometry distortion, or floaters.}
   \label{fig:comp-replica}
\end{figure*}

\begin{table*}[htb]
\footnotesize
\centering
\begin{tabular}{|l|ccc|ccc|ccc|c|c|c|}
\hline
\multicolumn{1}{|c|}{}                   & \multicolumn{3}{c|}{Mean}                                                                                                                 & \multicolumn{3}{c|}{Interpolation}                                                                                                        & \multicolumn{3}{c|}{Extrapolation}                                                                                                        &                                                                                           &                                                                            &                                                                           \\ \cline{2-10}
\multicolumn{1}{|c|}{\multirow{-2}{*}{}} & \multicolumn{1}{c|}{PSNR$\uparrow$}                          & \multicolumn{1}{c|}{SSIM$\uparrow$}                           & LPIPS$\downarrow$                          & \multicolumn{1}{c|}{PSNR$\uparrow$}                          & \multicolumn{1}{c|}{SSIM$\uparrow$}                           & LPIPS$\downarrow$                          & \multicolumn{1}{c|}{PSNR$\uparrow$}                          & \multicolumn{1}{c|}{SSIM$\uparrow$}                           & LPIPS$\downarrow$                          & \multirow{-2}{*}{\begin{tabular}[c]{@{}c@{}}w/ depth \\ input\end{tabular}} & \multirow{-2}{*}{\begin{tabular}[c]{@{}c@{}}training \\ time\end{tabular}} & \multirow{-2}{*}{\begin{tabular}[c]{@{}c@{}}render \\ (fps)\end{tabular}} \\ \hline
DVGO                                     & \multicolumn{1}{c|}{21.97}                         & \multicolumn{1}{c|}{0.781}                         & 0.487                         & \multicolumn{1}{c|}{27.36}                         & \multicolumn{1}{c|}{0.835}                         & 0.525                         & \multicolumn{1}{c|}{16.58}                         & \multicolumn{1}{c|}{0.727}                         & 0.448                         & 0                                                                                         & $\sim$10min                                                                & 0.4                                                                       \\ \hline
Plenoxels                                & \multicolumn{1}{c|}{27.54}                         & \multicolumn{1}{c|}{0.860}                         & 0.370                         & \multicolumn{1}{c|}{31.86}                         & \multicolumn{1}{c|}{0.903}                         & 0.333                         & \multicolumn{1}{c|}{23.22}                         & \multicolumn{1}{c|}{0.832}                         & 0.408                         & 0                                                                                         & $\sim$18min                                                                & 3748                                                                      \\ \hline
NeRF                             & \multicolumn{1}{c|}{30.89}                         & \multicolumn{1}{c|}{0.890}                         & 0.365                         & \multicolumn{1}{c|}{32.81}                         & \multicolumn{1}{c|}{0.904}                         & 0.358                         & \multicolumn{1}{c|}{28.98}                         & \multicolumn{1}{c|}{0.879}                         & 0.371                         & 0                                                                                         & $\sim$3h                                                                   & 0.03                                                                      \\ \hline
InstantNGP                               & \multicolumn{1}{c|}{31.44}                         & \multicolumn{1}{c|}{0.892}                         & 0.354                         & \multicolumn{1}{c|}{34.25}                         & \multicolumn{1}{c|}{0.917}                         & 0.323                         & \multicolumn{1}{c|}{28.63}                         & \multicolumn{1}{c|}{0.867}                         & 0.385                          & 0                                                                                         & $\sim$85s                                                                  & 9.25                                                                      \\ \hline
TSDF Fusion                              & \multicolumn{1}{c|}{27.64}                         & \multicolumn{1}{c|}{0.858}                         & 0.379                         & \multicolumn{1}{c|}{28.55}                         & \multicolumn{1}{c|}{0.869}                         & 0.371                         & \multicolumn{1}{c|}{26.73}                         & \multicolumn{1}{c|}{0.846}                         & 0.387                         & 1                                                                                         & -                                                                          & -                                                                         \\ \hline
DSNeRF                                   & \multicolumn{1}{c|}{31.16}                         & \multicolumn{1}{c|}{0.887}                         & 0.370                         & \multicolumn{1}{c|}{32.34}                         & \multicolumn{1}{c|}{0.895}                         & 0.369                         & \multicolumn{1}{c|}{29.99}                         & \multicolumn{1}{c|}{0.878}                         & 0.371                         & 1                                                                                         & $\sim$3h                                                                   & 0.03                                                                      \\ \hline
NeurMiPs                                 & \multicolumn{1}{c|}{33.29}                         & \multicolumn{1}{c|}{0.923}                         & 0.290                         & \multicolumn{1}{c|}{35.07}                         & \multicolumn{1}{c|}{0.938}                         & 0.271                         & \multicolumn{1}{c|}{31.52}                         & \multicolumn{1}{c|}{0.908}                         & \cellcolor[HTML]{FFD573}0.309                         & 1                                                                                         & $\sim$10h                                                                  & 0.73                                                                      \\ \hline   
NeRF-SLAM                                & \multicolumn{1}{c|}{\cellcolor[HTML]{FFD573}34.50} & \multicolumn{1}{c|}{\cellcolor[HTML]{FFD573}0.930} & \cellcolor[HTML]{FFD573}0.283 & \multicolumn{1}{c|}{\cellcolor[HTML]{FFD573}36.94} & \multicolumn{1}{c|}{\cellcolor[HTML]{FFD573}0.948} & \cellcolor[HTML]{FFD573}0.245 & \multicolumn{1}{c|}{\cellcolor[HTML]{FFD573}32.07} & \multicolumn{1}{c|}{\cellcolor[HTML]{FFD573}0.913} & 0.320 & 1                                                                                         & $\sim$88s                                                                  & 31.3                                                                      \\ \hline  

\textbf{PARF (ours)}                     & \multicolumn{1}{c|}{\cellcolor[HTML]{FFADB0}\textbf{35.09}} & \multicolumn{1}{c|}{\cellcolor[HTML]{FFADB0}\textbf{0.943}} & \cellcolor[HTML]{FFADB0}\textbf{0.228} & \multicolumn{1}{c|}{\cellcolor[HTML]{FFADB0}\textbf{37.00}} & \multicolumn{1}{c|}{\cellcolor[HTML]{FFADB0}\textbf{0.954}} & \cellcolor[HTML]{FFADB0}\textbf{0.204} & \multicolumn{1}{c|}{\cellcolor[HTML]{FFADB0}\textbf{33.18}} & \multicolumn{1}{c|}{\cellcolor[HTML]{FFADB0}\textbf{0.933}} & \cellcolor[HTML]{FFADB0}\textbf{0.253} & 1                                                                                         & \textbf{$\sim$40s}                                                         & 62.5                                                                      \\ \hline 

\end{tabular}
\caption{Quantitative evaluation results on the Replica dataset. Compared to baseline methods, PARF achieves the best performance in all three metrics and shows a significant boost in extrapolation ability.}
\label{tab:quantified_evaluation}
\end{table*}

\begin{figure*}[htb]
  \centering
   \includegraphics[width=0.9\linewidth]{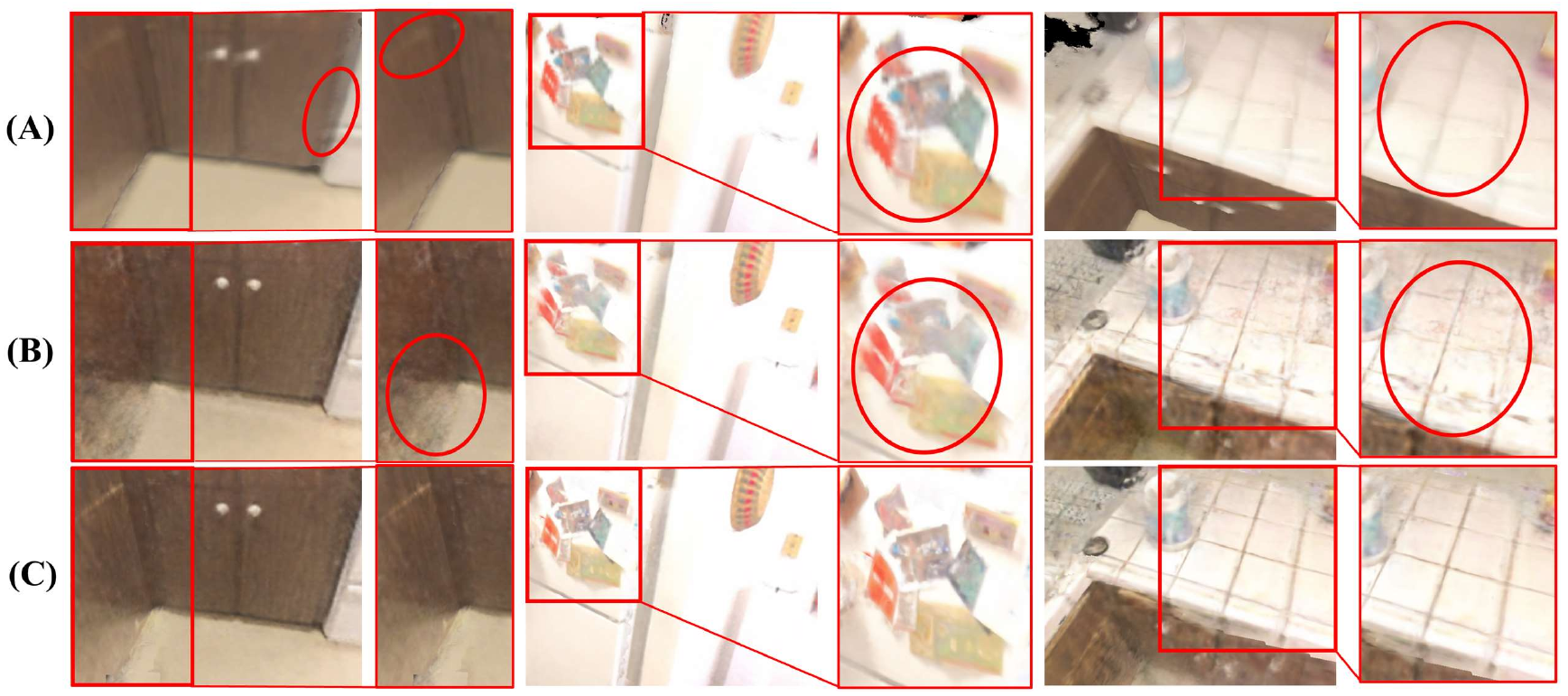}
   \caption{Qualitative comparison of TSDF Fusion(A), NeRF-SLAM(B), and PARF(C) on BundleFusion dataset. The views are rendered from extrapolation views that deviate significantly from the training views. TSDF Fusion and NeRF-SLAM suffer from blurry patterns and floaters, while PARF shows more robust rendering results.}
   \label{fig:comp-bf}
\end{figure*}

\section{Experiments}
In this section, we report the experimental results in detail. We first introduce the experiment settings, then show that PARF achieves high-quality render results and more robust view extrapolation compared to the SOTA methods.


\subsection{Datasets}
We perform experiments on the following public datasets: one synthetic dataset with ground truth depths and real-world datasets with noisy depths. 

\noindent\textbf{Replica~\cite{replica19arxiv}} consists of 18 scenes scanned and reconstructed from the real world, which can be rendered to RGB-D sequences. 
We conduct experiments on 8 sequences of Replica following existing method~\cite{rosinol2022nerf}. Specifically, we use 2000 frames with an interval of 10 frames for the training of each scene. Besides, 10 interpolation views and 10 extrapolation views are rendered as ground truth images for the evaluation of each scene. 

\noindent\textbf{BundleFusion~\cite{dai2017bundlefusion}} dataset includes real captured RGB-D sequences from 7 real-world indoor scenes. We obtain camera poses from the BundleFusion algorithm. For each scene, 2000 frames with an interval of 12 or 24 frames from each of 4 scenes (\texttt {apt0}, \texttt{apt2}, \texttt{copyroom}, and \texttt{office2}) are used for evaluation. 

\noindent\textbf{ScanNet~\cite{dai2017scannet}} We choose 3 scenes (\texttt {scene0012}, \texttt{scene0027}, and \texttt{scene0457}) from ScanNet dataset for evaluation. For each scene, 2000$\sim$5000 frames with an interval of 10 or 40 frames are chosen and the camera poses are also obtained from the BundleFusion algorithm~\cite{dai2017bundlefusion}.

Since no extra views can be obtained (like the Replica dataset), 
the BundleFusion and ScanNet datasets are only used to measure the interpolation ability in quantitative evaluation.

\subsection{Baselines}
We compare our method against the state-of-the-art methods for novel view synthesis, such as NeRF~\cite{mildenhall2020nerf}, DVGO~\cite{sun2022dvgo}, Plenoxels~\cite{fridovich2022plenoxels}, InstantNGP~\cite{muller2022instant}, which do not rely on depth input. Besides, methods like TSDF Fusion~\cite{zeng20163dmatch}, DS-NeRF~\cite{deng2022dsnerf}, NeurMiPs~\cite{lin2022neurmips}, NeRF-SLAM~\cite{rosinol2022nerf} that need geometry guidance are also compared.

We evaluate the standard \textbf{TSDF Fusion~\cite{zeng20163dmatch}} algorithm with the volume size of $512^3$ and the truncation band size of 10 voxels. 
For \textbf{DS-NeRF}~\cite{deng2022dsnerf}, we use dense depth maps as guidance instead of sparse point clouds to get better performance. 
\textbf{NeurMiPs}~\cite{lin2022neurmips} models the scene with multiple planar primitives. Since the plane parameters should be optimized with a global point cloud, it is difficult for NeurMiPs to be applied in an incremental reconstruction framework. We prepare the point clouds by fusing multi-view depth maps for the plane initialization stage.
During the evaluation of the mentioned baseline methods, we strictly follow the official hyper-parameters for a fair comparison.
\textbf{InstantNGP~\cite{muller2022instant}} is an extension of NeRF that enjoys fast convergence and rendering. We re-implement InstantNGP with pytorch-lightning and customized CUDA kernel functions~\cite{ngp_pl}. 
As a SLAM system, \textbf{NeRF-SLAM}~\cite{rosinol2022nerf} has a mapping stage that incorporates InstantNGP with a depth render loss. Since localization is out of the scope of this paper, only the mapping stage of NeRF-SLAM is evaluated. 
For a fair comparison with PARF, we follow the same hyper-parameter setting for networks and optimization when training InstantNGP and NeRF-SLAM.

To compare the convergence speed of PARF and NeRF-SLAM, we evaluate both interpolation and extrapolation quality with an appropriate iteration interval for each method, which is shown in Fig.~\ref{fig.teaser}(b). Note that we only evaluate the effectiveness of the proposed hybrid representation in Fig.~\ref{fig.teaser}(b), so we assume all the observations are available from the start of training. We further evaluate the efficiency under incremental reconstruction setup in Fig.~\ref{fig:increm}.

\subsection{Evaluation}
In this section, we report the results of quantitative and qualitative experiments on three datasets, which help validate the effectiveness of PARF in terms of rendering quality, convergence speed, as well as incremental reconstruction performance.

\noindent\textbf{Quantitative evaluation.}
We evaluate the interpolation and extrapolation performance of baselines and PARF on the Replica dataset in Tab.~\ref{tab:quantified_evaluation}.
By comparing baselines with and without depth input, it can be found that geometric guidance generally enables better render quality by reducing geometry ambiguity, especially for extrapolation views. 
By implementing geometric guidance in a hybrid representation, PARF significantly outperforms all the baselines for both interpolation and extrapolation settings. 
Note that the hybrid representation of PARF helps improve the extrapolation ability by a large margin. 

NeurMiPs shows worse performance than PARF because 
the plane-only representation is relatively hard to optimize and cannot accurately fit complex geometries (like chair legs and flowers). On the contrary, PARF enjoys a sound combination of primitive and non-primitive areas, which is more flexible and robust to the level of detail modeling.



\begin{table}[h]
    \small
    \centering
    \begin{tabular}{c | c | c c c}
        \specialrule{.08em}{0pt}{0pt}
        \textbf{Dataset}     &  \textbf{Methods}  & \textbf{PSNR}$\uparrow$ & \textbf{SSIM}$\uparrow$ & \textbf{LPIPS}$\downarrow$ \\
        \specialrule{.08em}{0pt}{0pt}
        \multirow{3}{*}{ScanNet} & InstantNGP& 21.04 & 0.685 & 0.530 \\
        {} & NeRF-SLAM & 23.28 & 0.716 & 0.490 \\
        {} & PARF      & \textbf{23.93} & \textbf{0.753} & \textbf{0.474} \\
        \specialrule{.05em}{0pt}{0pt}
        \multirow{3}{*}{BF} & InstantNGP & 21.42 & 0.724 & 0.460 \\
        {} & NeRF-SLAM & 24.98 & 0.749 & 0.394 \\
        {} & PARF      & \textbf{25.82} & \textbf{0.760} & \textbf{0.363} \\
        
        \specialrule{.08em}{0pt}{0pt}
    \end{tabular}
    \caption{Quantitative evaluation on ScanNet and BundleFusion(BF) datasets. 
    }
    \label{tab:evaluation_scannet}
\end{table}

From Tab.~\ref{tab:evaluation_scannet}, we find that even the real-world datasets include noisy depth maps and imperfect poses, PARF still consistently outperforms InstantNGP and NeRF-SLAM on all metrics under the interpolation views, which proves the effectiveness of the proposed hybrid representation and primitive-aware fusion framework.



\begin{figure}[htb] 
\centering
    \includegraphics[width=0.85\linewidth]{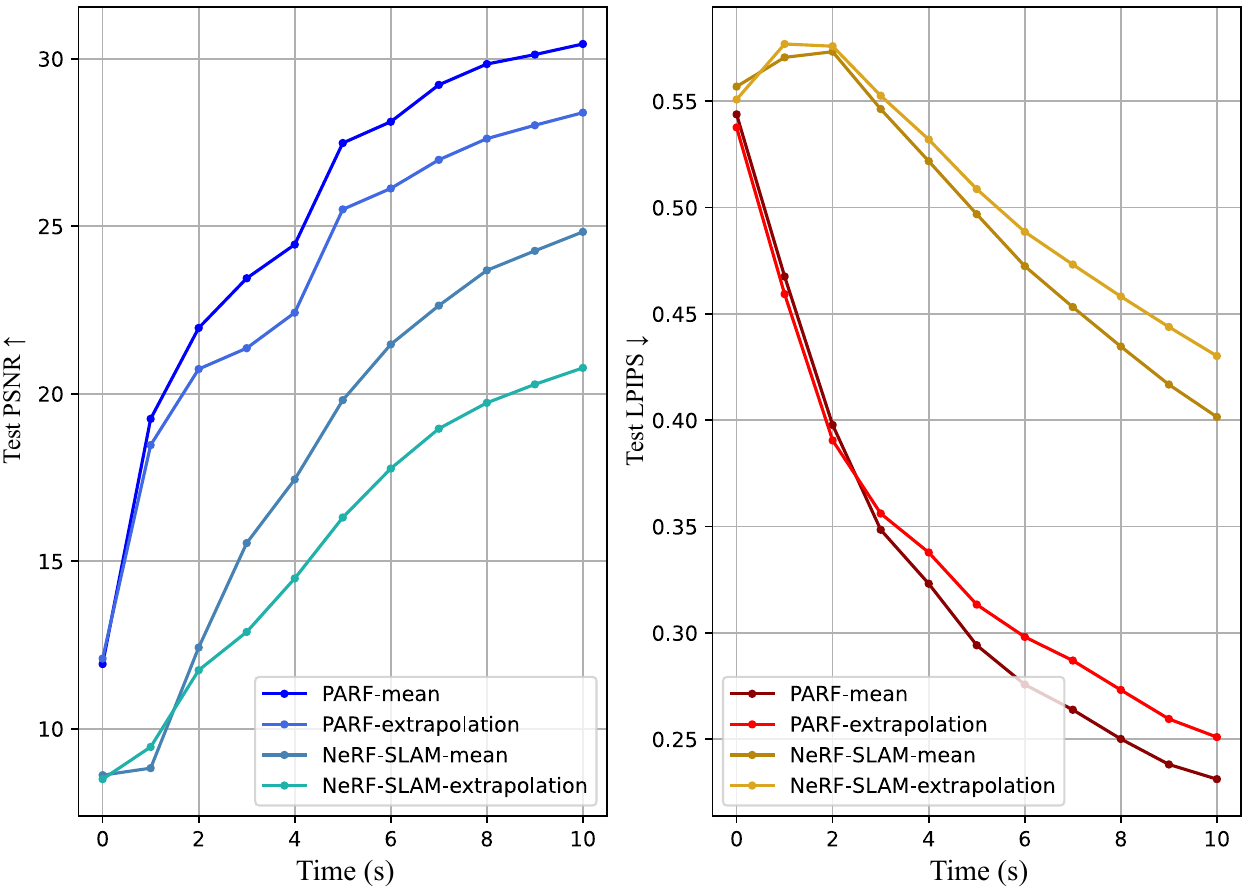} 
    \caption{Incremental reconstruction evaluation.}
    \label{fig:increm}
\end{figure}

\vspace{0.2cm}
\noindent\textbf{Qualitative evaluation.}
Fig.~\ref{fig:comp-replica} and Fig.~\ref{fig:comp-bf} show the visual comparison on different datasets. The rendering details shows that our method achieves the highest rendering quality and robustness. 
In both figures, TSDF Fusion shows blurry rendering results because of the simple multi-view color averaging strategy. Because of the insufficient scene regularization, DS-NeRF and NeRF-SLAM shows floaters in textureless regions and extrapolation views, which results in an apparent performance drop.
In Fig.~\ref{fig:comp-replica}, NeurMiPs suffers from distorted rendering results in the regions of complex geometry. Besides, NeurMips is very sensitive to plane initialization, which often results in holes due to the difficulty of plane parameter optimization. 
By comparison, the qualitative performance of PARF is more stable and accurate due to the advantage of the primitive-aware scene sensation and the hybrid representation.


\noindent\textbf{Speed Analysis.}
The \textbf{convergence speed} evaluation results are shown in Fig.~\ref{fig.teaser}(b). With the help of the effective combination with primitive-based rendering, PARF enjoys the highest converge speed compared to all the learning-based baselines that claimed for fast convergence, including Plenoxels, DVGO, InstantNGP, and NeRF-SLAM.

The \textbf{incremental reconstruction} performance is also evaluated. By comparing to NeRF-SLAM in Fig.~\ref{fig.teaser}(a) and Fig.~\ref{fig:increm}, PARF enjoys much faster convergence and can enable on-the-fly radiance fusion.

\subsection{Ablation Study}

\noindent\textbf{Observation sparsity.}
Primitives provide vital prior for scene perception, which enables robust performance even when the observation is relatively sparse. We evaluate PARF and NeRF-SLAM with different input sparsity on \texttt{office0} of Replica. The sparsity $n$ means we take one of every $n$ images from the original sequence (2000 frames) as input. The results are shown in Tab.~\ref{tab:ablation-sparsity-replica}, which demonstrate the robust performance of PARF even with very limited views ($<20$ views).

\begin{table}[htb]
\centering
\small
\begin{tabular}{c|c|cccc}
\hline
\multirow{2}{*}{\textbf{Methods}} & \multirow{2}{*}{\textbf{Metrics}}
& \multicolumn{4}{c}{\textbf{Sparsity}} \\
\cline{3-6}
&
& \textbf{20} & \textbf{60} & \textbf{100} & \textbf{140} \\
\hline
\multirow{3}{*}{NeRF-SLAM} & PSNR $\uparrow$ & 32.09 & 31.27 & 28.09 & 25.68 \\
 & SSIM $\uparrow$ & 0.919 & 0.908 & 0.879 & 0.831  \\
 & LPIPS $\downarrow$  & 0.274 & 0.287 & 0.316 & 0.366  \\
\hline
\multirow{3}{*}{{\makecell{PARF}}}
& PSNR $\uparrow$& \textbf{33.18} & \textbf{32.67} & \textbf{31.53} & \textbf{29.76} \\
& SSIM $\uparrow$ & \textbf{0.934} & \textbf{0.923} &  \textbf{0.911} & \textbf{0.905} \\
 & LPIPS $\downarrow$  & \textbf{0.192} & \textbf{0.201} & \textbf{0.204} & \textbf{0.214} \\
\hline
\end{tabular}
\caption{Ablation study of observation sparsity.}
\label{tab:ablation-sparsity-replica}
\end{table}

\noindent\textbf{Sampling Strategy.} 
We conducted an ablation study to validate the effectiveness of our primitive guided sampling strategy. 
Since depth maps are available, a more straightforward way is to sample with depth values. 
Specifically, if a ray holds a valid depth value on the depth image, we only sample the points located on and behind the depth value during training.
However, depth values from sensors inevitably contain noise, which is harmful to direct depth guidance. 
We add Gaussian noise with mean $0cm$ and sigma $100cm$ to depth images. The experimental results on Replica \texttt{office0} demonstrate the robustness of the hybrid sampling strategy of PARF, while direct depth guidance shows massive quality degradation given noisy depth input. 


\begin{table}[htb]
\small
\centering
\begin{tabular}{c|ccc}
\hline
\textbf{Guidance} & \textbf{PSNR}$\uparrow$ & \textbf{SSIM}$\uparrow$ & \textbf{LPIPS}$\downarrow$\\
\hline
depth (w/o noise) & 29.19 & 0.881 & 0.305 \\
depth (w/ noise) & 20.09 & 0.698  & 0.539\\
primitive (w/o noise) & \textbf{33.18} & \textbf{0.934} & \textbf{0.192}\\
primitive (w/ noise) & \textbf{33.00} & \textbf{0.933} & \textbf{0.196}\\
\hline
\end{tabular}
\caption{Ablation study of sampling strategy on Replica.}
\label{tab:ablation-depthnoise}
\end{table}



\section{Conclusion}
In this paper, we introduce PARF, a \textbf{P}rimitive-\textbf{A}ware \textbf{R}adiance \textbf{F}usion method for indoor scene radiance field reconstruction and editing. 
By combining volumetric and primitive rendering in a hybrid neural representation, we successfully merge semantic parsing, primitive extraction, and radiance fusion into a single framework. 
PARF achieves significant improvement in convergence speed, strong view extrapolation performance, and realistic semantic editing effects simultaneously. 
Since the discrete semantic volume may lead to jagged primitive boundaries for novel view synthesis,  
future work includes combining the semantic information in a more compact manner and adding more kinds of primitives for more effective reconstruction. 

\vspace{0.3cm}
\noindent\textbf{Acknowledgements} 
This work is supported 
in part by Natural Science Foundation of China (NSFC) under contract No.62125106, 61860206003, 62088102, 62171255, 
in part by Ministry of Science and Technology of China under contract No. 2021ZD0109901,
in part by Tsinghua-Toyota Joint Research Fund.

{\small
\bibliographystyle{ieee_fullname}
\bibliography{egbib}
}

\end{document}